\pdfoutput=1
\documentclass[11pt]{article}

\usepackage[]{acl}

\usepackage{times}
\usepackage{latexsym}
\usepackage{graphicx}
\usepackage{amsfonts}
\usepackage{indentfirst}
\usepackage{multirow}
\usepackage[T1]{fontenc}
\usepackage[utf8]{inputenc}

\usepackage{microtype}
\usepackage{amsmath}
\usepackage[marginal]{footmisc}

\title{Co-VQA : Answering by Interactive Sub Question Sequence}

\author{Ruonan Wang\textsuperscript{\textdagger}, \ Yuxi Qian\textsuperscript{\textdagger}, \ Fangxiang Feng\textsuperscript{\textdagger}, \ Xiaojie Wang\textsuperscript{\textdagger$\ast$}, \ Huixing Jiang\textsuperscript{$\ddagger$} \\
        \textsuperscript{\textdagger}Beijing University of Posts and Telecommunications \\ 
        \textsuperscript{$\ddagger$}Meituan-Dianping Group \\ 
        \textsuperscript{\textdagger}\texttt{\{wangrn, qianyuxi, fxfeng, xjwang\}@bupt.edu.cn} \\
        \textsuperscript{$\ddagger$}\texttt{jianghuixing@meituan.com} \\
        }
\begin{document}
\maketitle
\begin{abstract}
Most existing approaches to Visual Question Answering (VQA) answer questions directly, however, people usually decompose a complex question into a sequence of simple sub questions and finally obtain the answer to the original question after answering the sub question sequence(\textbf{SQS}). By simulating the process, this paper proposes a conversation-based VQA (\textbf{Co-VQA}) framework, which consists of three components: Questioner, Oracle, and Answerer. Questioner raises the sub questions using an extending HRED model, and Oracle answers them one-by-one. An \textbf{Adaptive Chain Visual Reasoning Model (ACVRM)} for Answerer is also proposed, where the question-answer pair is used to update the visual representation sequentially. To perform supervised learning for each model, we introduce a well-designed method to build a SQS for each question on VQA 2.0 and VQA-CP v2 datasets. Experimental results show that our method achieves state-of-the-art on VQA-CP v2. Further analyses show that SQSs help build direct semantic connections between questions and images, provide
question-adaptive variable-length reasoning chains, and with explicit interpretability as well as error traceability.
\end{abstract}

\section{Introduction}
\noindent
Visual Question Answering \cite{Agrawal2015VQAVQ} requires to answer questions about images. It has to process visual and language information simultaneously, which is a basic ability of advanced agents. \footnote{\textsuperscript{$\ast$}Xiaojie Wang is the corresponding author.}Therefore, it has attracted more and more attention \cite{Anderson2018BottomUpAT, Lu2016HierarchicalQC, Goyal2017MakingTV, Agrawal2018DontJA}. The conventional approach \cite{Agrawal2015VQAVQ} for Visual Question Answering (VQA) is to encode image and question separately and incorporate the representation of each modality into a joint representation. Recently, with the proposal of Transformer \citep{Vaswani2017AttentionIA}, based on previous dense co-attention models \citep{Kim2018BilinearAN, Nguyen2018ImprovedFO}, some methods \citep{Yu2019DeepMC, Gao2019DynamicFW} further adopt self-attention mechanism to exploit the fine-grained information in both visual and textual modalities. Meanwhile, to enrich indicative information about the image contained in the visual representation, some researchers \citep{Cadne2019MURELMR, Li2019RelationAwareGA} have explored different methods of relational reasoning to capture the relationship between objects.

\begin{figure}[t]
\begin{center}
\includegraphics[width=1.0\linewidth]{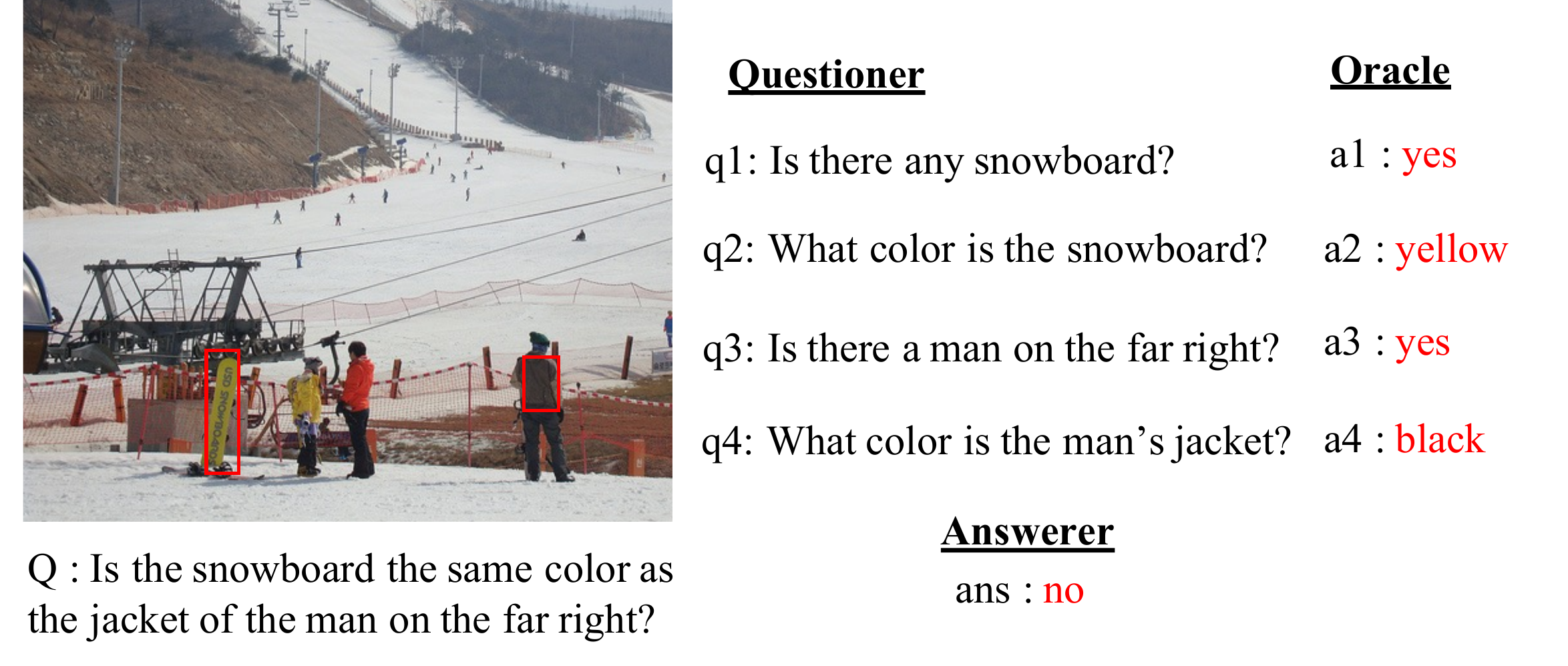}
\end{center}
\caption{An illustrative example. After a sequence of four sub questions and their answers 
\{(q1,a1),(q2,a2),(q3,a3),(q4,a4)\}, its easier to answer the original question.}
\label{fig:SQS}
\end{figure}

Though above methods have achieved significantly improved performances on real datasets \cite{Agrawal2015VQAVQ,Goyal2017MakingTV}, there are still some issues unsolvable. Most existing approaches answer questions directly, however, it is often difficult, especially to answer complex questions. On the one hand, achieving holistic scene understanding in one round is pretty challenging. On the other hand, performing the whole Q\&A process in one round lacks interpretability, and it is difficult to locate the errors when the model runs into wrong answers. To address the above difficulties, motivated by theory of mind \citep{Leslie1987PretenseAR}, as shown in Figure~\ref{fig:SQS}, we imagine an internal conversation for answering the original question, where a sub question sequence (\textbf{SQS}, which includes several simple sub questions, we use \textbf{SQ} to refer to sub question later) is raised and answered one-by-one progressively. Finally, the answer to the original question is obtained by capturing joint information accumulated in the whole SQS. This way has several significant cognitive advantages: 1) SQSs with different numbers of sub questions will be automatically generated for different questions, resulting in question-adaptive variable-length reasoning chains, 2) a SQS gives a clear reasoning path, it therefore provides explicit interpretability and traceability of errors, 3) different questions are likely to contain the same SQs or SQSs, these common SQs even SQSs help improve the generalization ability of models, 4) SQs are usually more simple and directly related to the images, which help to strengthen the semantic connections between linguistic and visual information.

To achieve above advantages, we therefore propose a \textbf{Conversation-based VQA (Co-VQA}) framework which includes an internal conversation for VQA. It consists of three components: \textbf{Questioner}, \textbf{Oracle} and \textbf{Answerer}. As shown in Figure~\ref{fig:SQS}, once a question is raised, Questioner asks some SQs, and Oracle provides answers one-by-one. Their conversation brings a SQS and the corresponding answer sequence. When there is no more SQ to be generated, the internal conversation is finished and Answerer gives the final answer to the original question.

Questioner employs the hierarchical recurrent encoder-decoder architecture \cite{Sordoni2015AHR}, and we adopt a representative VQA model \cite{Anderson2018BottomUpAT} as Oracle. For Answerer, we propose an \textbf{Adaptive Chain Visual Reasoning Model (ACVRM)} to accomplish an explicit progressive reasoning process based on SQS, where SQs are used to guide the update of visual features by an extended graph attention network \cite{Velickovic2018GraphAN} gradually. Meanwhile, the answers of SQs are utilized as additional supervision signals to guide the learning process. Further, to provide supervision information for the above three models during training, we propose a well-designed method to construct a SQS for each question, which is based on linguistic rules and natural language processing technology. VQA-SQS and VQA-CP-SQS datasets are obtained after applying this method to VQA 2.0 \cite{Goyal2017MakingTV} and VQA-CP v2 \cite{Agrawal2018DontJA} datasets.

Our contributions can be concluded into three-fold:
\begin{itemize}
\item We introduce a Conversation-based VQA (Co-VQA) framework, which consists of three components: Questioner, Oracle and Answerer. The frame is different from existing VQA methods in principle. 
\item  An Adaptive Chain Visual Reasoning Model (ACVRM) for Answerer is proposed, where the question-answer pair is used to update visual representation sequentially.
\item Co-VQA achieves the new state-of-the-art performance on the challanging VQA-CP v2 dataset. Moreover, SQSs help to build direct semantic connections between questions and images, they provide question-adaptive variable-length reasoning chains with explicit interpretability as well as error traceability.
\end{itemize}

\section{Related Work}
\noindent
\textbf{Visual Question Answering.} The current dominant framework for VQA consists of an image encoder, a question encoder, multimodal fusion, and an answer predictor \cite{Agrawal2015VQAVQ}. To avoid the noises caused by global features, methods\cite{Yang2016StackedAN, Malinowski2018LearningVQ} introduce various image attention mechanisms into VQA. Instead of directly using visual features from  CNN-based feature extractors, to improve the performance of model, BUTD\cite{Anderson2018BottomUpAT} adopts Faster R-CNN \cite{Ren2015FasterRT} to obtain candidate regional features while Pythia\cite{Jiang2018PythiaVT} integrates the regional feature with grid-level features. Meanwhile, \citet{Lu2016HierarchicalQC, Nam2017DualAN} put more attention on learning better question representations. To merge information from different modalities sufficiently, MFB\citep{Yu2017MultimodalFB} and MUTAN\citep{Benyounes2017MUTANMT} explored higher-order fusion methods. Further, BAN\cite{Kim2018BilinearAN} and DCN\cite{ Nguyen2018ImprovedFO} propose dense co-attention model which directly establish interaction between different modalities with word-level and regional features. Moreover, with the proposal of Transformer \citep{Vaswani2017AttentionIA}, MCAN \citep{Yu2019DeepMC} and DFAF \citep{Gao2019DynamicFW} adopt self-attention mechanism to fully excavate the fine-grained information contained in text and image. Meanwhile, to fully cover the holistic scene in an image, MuREL \cite{Cadne2019MURELMR} and ReGAT \cite{Li2019RelationAwareGA} explicitly incorporate relations between regions into the interaction process.

\citet{Selvaraju2020SQuINTingAV} also proposed sub questions but with very different motivation and methods. They found consistency issues in current VQA models which answer the reasoning questions correctly but fail on associated low-level perception questions. They therefore construct independent perception questions that serve as SQs to answer the reasoning questions, and proposed SQuINT to force a VQA model to attend to the same regions when answering the reasoning questions and their associated Perception SQ. The dataset proposed in this paper is different from them because our model needs a sequence of SQs to form a visual dialogue.
\noindent
\paragraph{Visual Dialogue.} Different from VQA, Visual dialogue (VD) is a continuous conversation for images. Several VD tasks (Visual Dialog \citep{Das2017VisualD}, GuessWhich \citep{Chattopadhyay2017EvaluatingVC}, GuessWhat?! \citep{Vries2017GuessWhatVO}, MMD \citep{Saha2018TowardsBL}) have been proposed. GuessWhat?!, as a goal-directed dialogue task, requires both players to continuously clarify the reference object through dialogue. The Oracle provides the Questioner with relevant information about the target object by constantly answering yes/no questions raised by the Questioner, and the Guesser generates the final answer based on the historical dialogue. Following the setting, our Co-VQA framework consists of three components, in which Questioner raises SQs, and Oracle answers them one-by-one, finally, Answerer obtains the answer to the original question.

\begin{figure}[t]
    \begin{center}
    \includegraphics[width=1.0\linewidth]{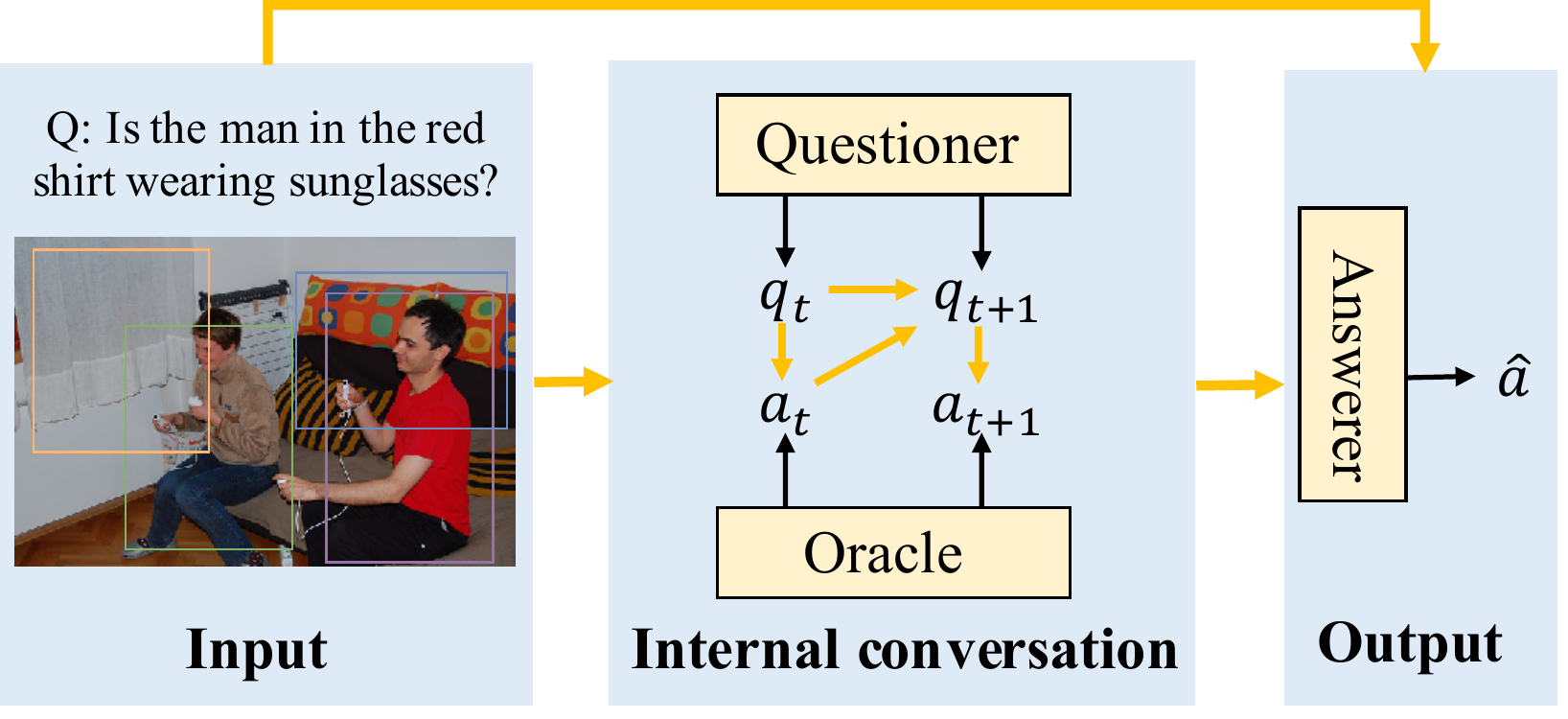}
    \end{center}
    \setlength{\belowcaptionskip}{-0.4cm} \caption{Overall illustration and data flow structure diagram of Co-VQA framework.}
    \label{fig:covqa}
\end{figure}

\begin{figure}[t]
    \begin{center}
    \includegraphics[width=1.0\linewidth]{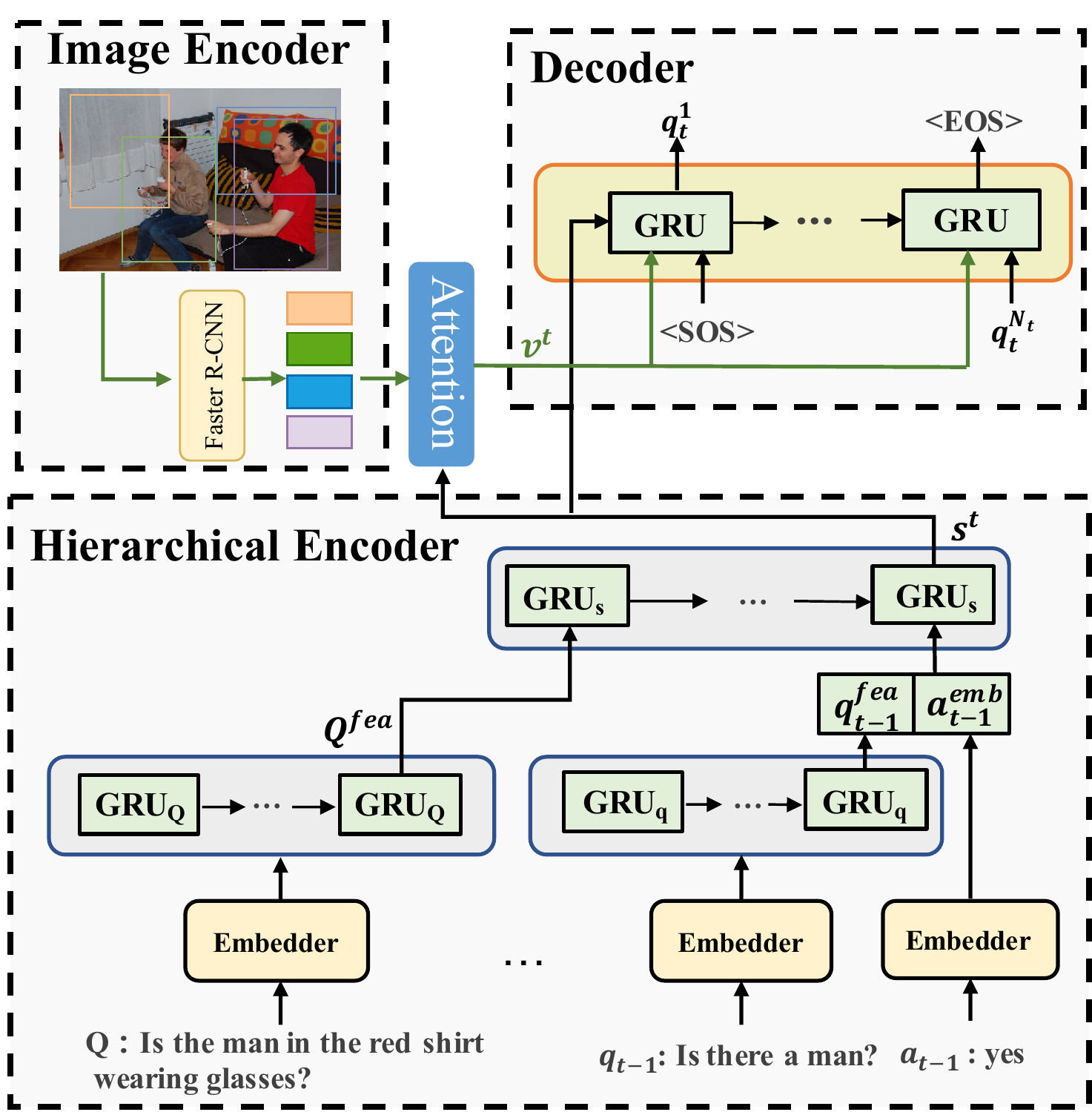}
    \end{center}
    \caption{Overview of the \textbf{Questioner} model which is based on extending HRED model. There are three modules: Image Encoder, Hierarchical Encoder, Decoder.}
    \label{fig:Questioner}
\end{figure}
\begin{figure*}[t]
    \begin{center}
    \includegraphics[width=1.0\linewidth]{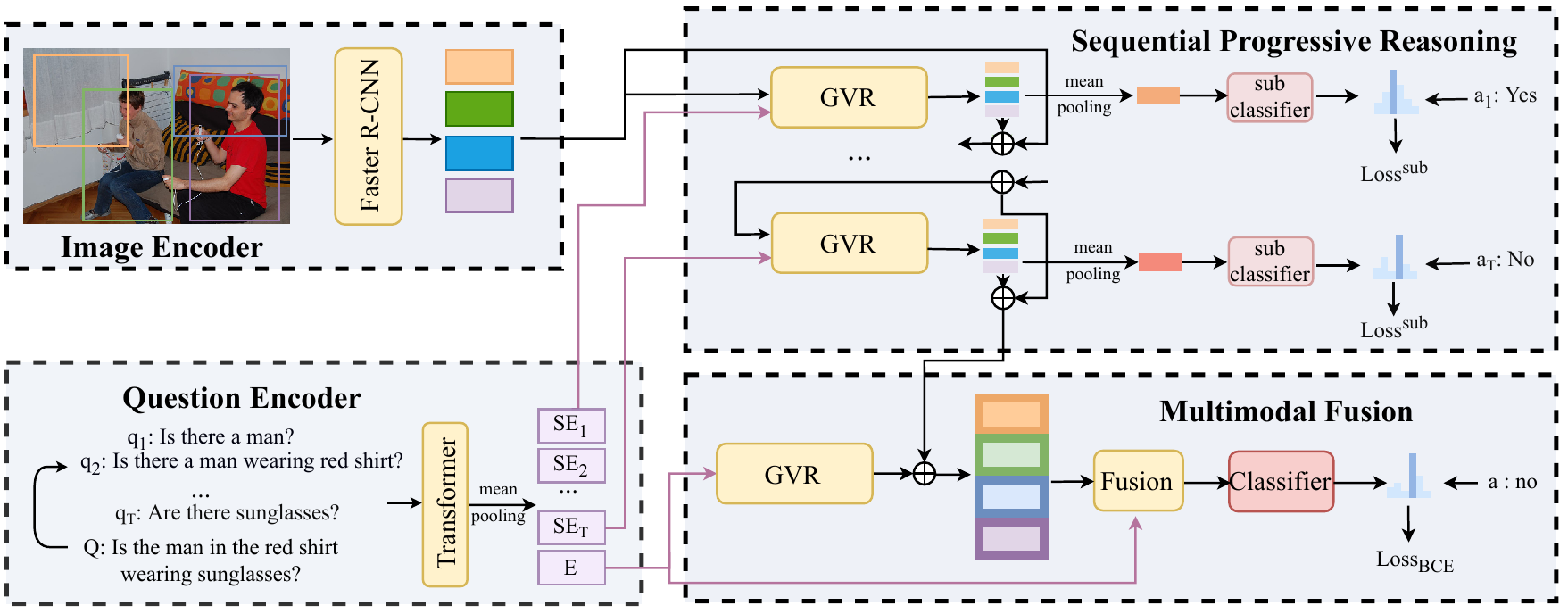}
    \end{center}
    \caption{Model architecture of the proposed \textbf{ACVRM} for Answerer. There are four functional modules: Image Encoder, Question Encoder, Sequential Progressive Reasoning and Multimodal Fusion.}
    \label{fig:ACVRM}
\end{figure*}

\section{Approach}
\noindent
Figure~\ref{fig:covqa} shows the overall structure and data flows of Co-VQA, where the Questioner, the Oracle, and the Answerer are three major components. Given an input image $I$ and a question $Q$, Co-VQA aims to predict the correct answer from the candidate answer set $A^{\ast}$. Specifically, the Questioner generates a new SQ $q_{t}$ for the next round by combining the information in $Q$, $I$ and the dialogue history $H_{t-1}=\{(q_1,a_1), \cdots, (q_{t-1}, a_{t-1})\}$. Then, Oracle produces appropriate answer $a_{t}$ for $q_{t}$. After accomplishing the last round of sub question-answer pair, Answerer utilizes the historical information accumulated throughout the process to obtain the final answer. In this section, we will introduce the three components in Section 3.1-3.3.

\subsection{Questioner}
\noindent
At round $t$, given an image $I$, a question $Q$ and the dialogue history $H_{t-1}=\{(q_1,a_1), \cdots, (q_{t-1}, a_{t-1})\}$, Questioner aims to generate a new SQ $q_{t}$, which could be denoted as:
\begin{equation}
q_t \sim P_{\theta_{Q}}(q | Q, I, H_{t-1}),
\end{equation}
where $\theta_{Q}$ denotes the parameters of Questioner. Generally, we build Questioner based on an extending hierarchical recurrent encoder decoder (HRED) architecture \cite{Sordoni2015AHR}. The overall structure of Questioner is depicted in Figure~\ref{fig:Questioner}.
\noindent
\paragraph{Image Encoder.}
Following common practice\cite{Anderson2018BottomUpAT}, we extract regional visual features from $I$ in a bottom-up manner by using Faster R-CNN model\cite{Ren2015FasterRT}. Each image will be encoded as a series of M regional visual features $R \in \mathbb{R}^{{M} \times 2048}$ with their bounding box $b = [x, y, w, h] \in \mathbb{R}^{{M} \times 4}$ ($M \in [10, 100]$ in our experiments).
\noindent
\paragraph{Hierarchical Encoder.} Embedding matrix Embedder is adopted to map $Q$ and each pair $(q_{i},a_{i})$ in $H_{t-1}$ to $Q^{emb}$ and $(q^{emb}_{i}, a^{emb}_{i})$  respectively. Then, two question-level encoder GRU, $GRU_Q$ and $GRU_q$, are deployed to obtain corresponding question feature $Q^{fea}$ and $q^{fea}_i$ for $Q$ and $q_i$.

$Q^{fea}$ is utilized as the first step input of session-level encoder GRU, $GRU_{s}$ to grasp global information of original question. $q^{fea}_{i}$ and $a^{emb}_{i}$ are concatenated as $qa^{fea}_{i}$, which is regarded as representation for sub question-answer pair. Meanwhile, it is treated as the i+1-th step input of $GRU_{s}$ to obtain context feature $s_{i+1}$:
\begin{equation}
     s_{i+1} = GRU_{s}([q^{fea}_i \hspace{0.2em} || \hspace{0.2em} a^{emb}_i], s_{i}),
\end{equation}
where $||$ represents concatenation. After encoding $H_{t-1}$, we obtain current context representation $s_{t}$.
\noindent
\paragraph{Decoder.} At decoding $q_{t}$, we employ an extra one-layer GRU as decoder, which is initialized by $s_{t}$. Then a question-guided attention is deployed to regional features $R$ to obtain the weighted visual feature $v_{t}$. Further, we fuse $v_{t}$ with $Embedder(q^i_{t})$ as the input of decoder at every time step $i$.

The negative log-likelihood loss is used for training, where T is the maximum round of dialogues:
\begin{equation}
    L({\theta}_Q) = -\sum_{t=1}^T log{P(q_t|Q, I, H_{t-1})}.
\end{equation}
\subsection{Oracle}
\noindent
The Oracle aims to constantly answer SQs raised by Questioner. Specifically, at round t, Oracle supplies the answer $a_t$ for SQ $q_t$, based on the image I and SQ $q_t$. We regard Oracle as a conventional VQA task and adopt the BUTD \cite{Anderson2018BottomUpAT}, which is a representative VQA method, as our Oracle.
\subsection{Answerer}
\noindent
Given a question $Q$, an image $I$ and a complete dialogue history $H_T = \{q_1,a_1,...,q_T,a_T\}$, the assignment of Answerer is to find out the most accurate $\hat{a}$ in the candidate answer set $A^{\ast}$, which could be denoted as:
\begin{equation}
     \hat{a} =\mathop{argmax}\limits_{{a} \in {A^{\ast}}}P_{\theta}(a|I,Q,H_T),
\end{equation}
\noindent
where $\theta$ denotes the parameters of Answerer. To accomplish this task, we propose an \textbf{Adaptive Chain Visual Reasoning Model (ACVRM)}, which consists of four components: Image Encoder, Question Encoder, Sequential Progressive Reasoning, and Multimodal Fusion. The overall structure of ACVRM is illustrated as Figure~\ref{fig:ACVRM}.

\subsubsection{Image and Question Encoder}
\noindent
Feature extraction modules are shown in the left part of Figure~\ref{fig:ACVRM}. Image encoder is the same as Questioner. For question encoder, we adopt a bidirectional Transformer \cite{Vaswani2017AttentionIA}. $Q$ and each SQ in $H_T$ will be padded to a maximum length and be encoded by bidirectional Transformer with random initialization, at last the corresponding question features $E \in \mathbb{R}^{d_q}, {\{SE_i\}}_{i=1}^T \in \mathbb{R} ^{{T} \times {d_q}}$ are obtained after mean pooling. To align the feature dimensions, we linearly map image feature $R$ to $V_0 \in \mathbb{R}^{{M} \times d_v}$. We set $d_q = d_v = 768$.

 \subsubsection{Sequential Progressive Reasoning (SPR)}
\noindent
\paragraph{Overall.} To realize progressive visual reasoning under the guidance of SQS, we utilize \textbf{Graph Visual Reasoning (GVR)} module, which will be introduced later, to gradually guide the update of visual features. Specifically, for $Q$ containing T SQs, the t-th step of SPR can be expressed as:
\begin{equation}
       V_{t}^{R} = GVR(V_{t-1},{SE}_t; \theta_{G}),
\end{equation}
where $V_{t}^{R}$ represents the t-th step visual feature, and $\theta_{G}$ denotes parameters for GVR. Then, the residual connection is deployed in each round to preserve historical information and avoid vanishing gradients. Therefore, the updated visual feature for the t-th round can further be depicted as:
\begin{equation}
    V_t= V_{t-1}+ V_{t}^{R}. 
\end{equation}

Furthermore, each $q_{t}$ has a corresponding answer $a_{t}$, which supplies an additional supervision signal for training. For each step t, we adopt a shared two-layer MLP as the sub classifier and then utilize average $V_{t}^{R}$ as input. A cross-entropy loss is used for classification, which is denoted as $Loss_t^{sub}$.

\begin{figure}[t]
    \begin{center}
    \includegraphics[width=1.0\linewidth]{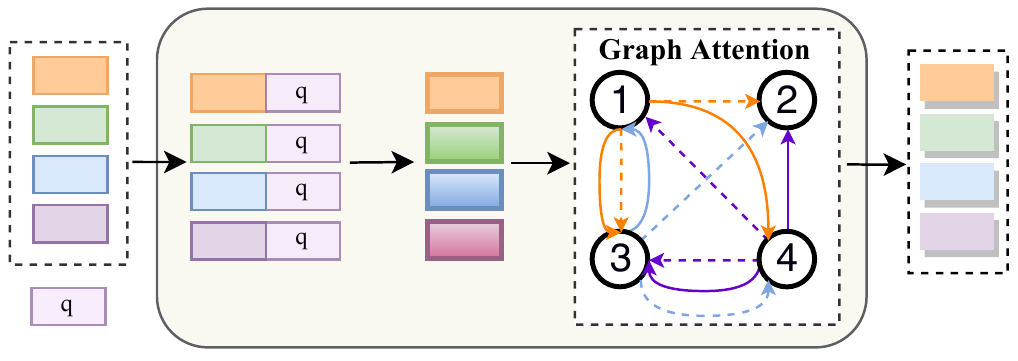}
    \end{center}
    \setlength{\belowcaptionskip}{-0.4cm}\caption{Flowchart of the \textbf{GVR}, including two parts: multimodal fusion based on concatenation and relation reasonsing based graph attention network.}
    \label{fig:gvr}
\end{figure}

\noindent
\paragraph{Graph Visual Reasoning.} Inspired by ReGAT \cite{Li2019RelationAwareGA}, we utilize an extended Graph Attention Network \cite{Velickovic2018GraphAN} to learn relations between objects. An overall illustration of GVR is shown in Figure~\ref{fig:gvr}. The whole reasoning process is abbreviated as $V^R = GVR(V, q)$, which consists of two parts: feature fusion and relational reasoning.

At first, the question representation $q$ is concatenated with each of the $M$ visual features $v_i$, which we write as $[v_i \hspace{0.2em} || \hspace{0.2em} q]$, then we compute a joint embedding as:
 \begin{equation}
     v^{'}_{i} = W([v_i \hspace{0.2em} || \hspace{0.2em} q]) \hspace{1em} for \hspace{0.5em}  i = 1,...,M,
 \end{equation}
where $W\in \mathbb{R}^{d_q \times {(d_q+d_v)}}$, and $v^{'}_{i} \in \mathbb{R}^{d_q}$ is conducted as the initial value of node in the graph $G(V,E)$, where $e_{i j}$ denotes edges between nodes. Then, to reduce the interference caused by irrelevant information, we design a masked multi-head attention for relational reasoning. Specially, for each head, inspired by \citet{Hu2018RelationNF}, the attention weight not only depends on visual-feature weight $\alpha^{h,v}_{i j}$ , but also bounding-box weight $\alpha^{h,b}_{i j}$, we formulate non-normalized attention weight $e_{i j}$ as:
\begin{gather}
    e^{h}_{i j}= \alpha_{i j}^{h, v} + \log(\alpha^{h, b}_{i j}), \\
    \alpha_{i j}^{h, v} = \frac{({W_{q}^h v^{'}_{i}})^{T} \cdot {W_{k}^h v^{'}_{j}}}{\sqrt{d_h}}, \\
    \alpha_{i j}^{h, b} = \max\left\{0, w\cdot f_{b}(b_{i}, b_{j})\right\},
\end{gather}
where $d_{h}=\frac{d_q}{H}$, $H$ denotes the number of head and we set $H=8$, $W_{q}^h \in \mathbb{R}^{d_h \times d_q}$, $W_{k}^h \in \mathbb{R}^{d_h \times d_q}$, $f_b(\cdot , \cdot)$ first computes relative geometry feature $(log(\frac{|x_i{-}x_j|}{w_i}), log(\frac{|y_i{-}y_j|}{h_i}), log(\frac{w_j}{w_i}), log(\frac{h_j}{h_i}))$, then embeds it into a $d_h$-dimensional feature by computing cosine and sine functions of different wavelengths, $w \in \mathbb{R}^{d_h}$. Furthermore, according to $e^{h}_{i j}$, to learn a sparse neighbourhood $N^{h}_{i}$ for each node i, we adopt a ranking strategy as $N^{h}_{i} = top_K(e^{h}_{i j})$, where $top_K$ returns the indices of the K largest values of an input vector, and we set K=15.

By employing above mechanism, output features of each head are concatenated, where $W^{h}_{v} \in \mathbb{R}^{d_h \times d_q}$:
\begin{equation}
    v_i^R={||}_{h=1}^{H}\sigma(\sum_{ j \in N^{h}_{i}} softmax(e_{i j}^{h}) \cdot {W_v^h v_j^{'}} ).
\end{equation}

\subsubsection{Fusion Module}
\noindent
SQ-aware visual features $V_T$ are obtained after completing the whole process of $SPR$. To sufficiently integrate the information of two modalities, we utilize $Q$ to convert $V_T$ into final context-aware $\Tilde{V}$ through $GVR$:
\begin{equation}
    \Tilde{V} = GVR(V_T, E).
\end{equation}

\begin{table*}
    \centering
    \resizebox{0.7\linewidth}{!}{
    \begin{tabular}{l|c|c|c|c|c}
    \hline
    \multicolumn{5}{c|}{Validation} & Test-std \\
    \hline
    Model   & All & Y/N & Num & Other & All \\
    \hline
    Bottom-Up     & 63.37        & 80.4   & 43.02  & 55.96 & 65.67\\
    BAN      & 66.04   &   -  & - & - & - \\
    MuREL    & 65.14   & - & - & - & 68.41 \\
    ReGAT$\ast$ & 67.18 & - & - & - & 70.58$\dagger$\\
    DFAF     & 66.66 & - & - &  - & 70.34$\dagger$ \\
    MCAN     & 67.2(67.14$\pm{0.04}\ddagger$) & 84.82$\ddagger$  & 49.24\textbf{$\ddagger$}  & 58.44$\ddagger$ & 70.9$\dagger$ \\
    MLIN     & 66.53   & - & - & - & 70.28$\dagger$\\
    \hline
    Ours     & \textbf{67.26$\pm{0.02}$} & 84.71  & \textbf{50.38}  & 58.44 & 70.39\\
    \hline
    \end{tabular}}
    \centering
	\caption{Performance on VQA 2.0 validation split and test-standard splits. "$\ast$" means ensembling result.  "$\dagger$" means training with augmented VQA samples from Visual Genome. "$\ddagger$" based on our re-implementations.} 
	\label{tab:vqaval}
\end{table*}

Then, we employ the same multi-modal fusion strategy as \citet{Anderson2018BottomUpAT} to obtain a joint representation $H$. For Answer Predictor, we adopt a two-layer multi-layer perceptron (MLP) as classifier, with $H$ as the input. Binary cross entropy is used as the loss function. Thus, final loss can be formulated as:
\begin{equation}
    Loss = Loss_{BCE} + \sum_{t=1}^TLoss_t^{sub}.
\end{equation}

\section{Experiments}
%In this section, we conduct experiments to evaluate the performance of our Co-vqa on the %largest VQA
%benchmark dataset, VQA-v2\cite{Goyal2017MakingTV} and VQA-CP v2 datasets. Since the length %of SQS,sub loss, and the coherence of SQS may influence final performance, we perform %extensive quantitative and
%qualitative ablation studies to explore the potential factors affecting of the model.
\subsection{Datasets}
\noindent
We evaluate our approach on two widely used datasets:

\textbf{VQA 2.0} \cite{Goyal2017MakingTV} is composed of real images from MSCOCO \cite{Lin2014MicrosoftCC} with the same train/validation/test splits. For each image,  an average of 3 questions are generated. These questions are divided into 3 categories: Y/N, Number, and Other. 10 answers are collected for each image-question pair from human annotators. The model is trained on the train set, but when testing on the test set, both train and validation set are used for training, and the max-probable answer is selected as the predicted answer.

\textbf{VQA-CP v2} \cite{Agrawal2018DontJA} is a derivation of VQA 2.0. In particular, the distribution of answers concerning to question types is designed to be different between train and test splits, which is aimed at overcoming language priors.

\paragraph{Construction of SQS dataset.} To provide the corresponding supervised signal for training Questioner, Oracle, and Answerer, we propose a well-designed method, which is chiefly based on linguistic rules and natural language processing technology. \textbf{VQA-SQS} and \textbf{VQA-CP-SQS} are obtained by applying this method on VQA 2.0 and VQA-CP v2 datasets. The details of the construction process and the specific statistical information of the two datasets can be found in Appendix.
\subsection{Implementation Details}
\noindent
\textbf{Training and inference.} During training, Questioner, Oracle, and Answerer are trained independently. For inference, given a question $Q$ and an image $I$, SQS is firstly generated through the cooperation between Questioner and Oracle, then $Q$, $I$ and the complete SQS is combined as the input of Answerer, and obtain the final answer.

\textbf{Parameters.} Each question is tokenized and padded with 0 to a maximum length of 14. For Questioner and Oracle, each word is embedded using 300-dimensional word embeddings. The dimension of the hidden layer in GRU is set as 1,024(except for $GRU_Q$ and $GRU_s$ with 1,324).

Our model is implemented based on PyTorch\cite{Paszke2017AutomaticDI}. In experiments, we use Adamax optimizer for training, with the mini-batch size as 256. For choice of the learning rate, we employ the warm-up strategy\cite{Goyal2017AccurateLM}. Specifically, we begin with a learning rate of 5e-4, linearly increasing it at each epoch till it reaches 2e-3 at epoch 4. After 14 epochs, the learning rate is decreased by 0.2 for every 2 epochs up to 18 epochs. We also adopt an early stopping strategy. For the transformer encoder, we fix the learning rate as 5e-5. Every linear mapping is regularized by weight normalization and dropout (p = 0.2 except for the classifier with 0.5).

\subsection{Results}
\noindent
To compare with existing VQA methods, we conduct several experiments to evaluate the performance of our Co-VQA framework, further, to verify the generation quality of the SQs and their impact on the performance of the overall model, Questioner and Oracle are tested additionally.

In Table~\ref{tab:vqaval}, we compare our method with previous work on VQA 2.0 validation and test-standard split. From Table~\ref{tab:vqaval}, it can be seen that on validation split, Co-VQA achieves the top-tier performance, an accuracy of 67.26, which surpasses that of  MCAN\cite{Yu2019DeepMC} by 0.06. Although the absolute improvement is slight, we report the standard deviation in Table~\ref{tab:vqaval}, compared with MCAN, the p-value is 0.006, so the improvements are statistically significant with p < 0.05. Moreover, we achieve an obvious performance improvement on the number questions. On VQA 2.0 test-standard split, without additional augmented samples from Visual Genome \cite{krishna2017visual}, our performance is still the third place. We assume the gap between the two splits is mainly due to the difference in SQS generation quality.

\begin{table}
    \centering
    \resizebox{\linewidth}{!}{
    \begin{tabular}{l|cccc}
    \hline
    Model & All & Y/N & Num & Other \\
    \hline
    MuREL & 39.54 & 42.85 &  13.17 & 45.04 \\
    ReGAT$\ast$ & 40.42 & - & - & - \\
    MCAN$\ddagger$  & 42.35 & 42.29 & 14.51 & \textbf{50.02} \\
    \hline
    Ours & \textbf{42.52} & \textbf{44.42} & \textbf{14.68} & 49.17 \\
    \hline
    \end{tabular}}
    \vspace{0.1cm}\caption{State-of-the-art comparison on the VQA-CP v2 dataset. "$\ast$" means ensembling result. "$\ddagger$" Results based on our re-implementations.}
    \label{tab:vqacp}
\end{table}

To demonstrate the generalizability of Co-VQA, we also conduct experiments on the VQA-CP v2 dataset, where the distributions of the train and test splits are quite different. Table~\ref{tab:vqacp} illustrates the overall performance, and our model gains a significant advantage (+2.1) over ReGAT. Compared with MCAN, our model also improved by 0.16.

For Questioner and Oracle, we train and evaluate the train/validation split of the VQA-SQS dataset.
\paragraph{Oracle.} The accuracy of Oracle is 93.73 and the average F-value is 90.13. On the one hand, the high accuracy is due to SQ itself being simple; On the other hand, decomposition of question leads to many same SQs, strengthening image-language correlation ability at SQ level.
\paragraph{Questioner.} For Questioner, the BLEU score is adopted to measure the quality of the generated SQs. As is shown in Table~\ref{tab:bleu}, we attribute the low BLEU scores to the diversity of syntax details.

\begin{table}
    \centering
    \begin{tabular}{ccc}
    \hline
    \textbf{BLEU-1} & \textbf{BLEU-2} & \textbf{BLEU-3}\\
    \hline
    67.8 & 38.4 & 21.2  \\
    \hline
    \end{tabular}
    \caption{BLEU evaluation scores of Questioner. We don't report BLEU-4 score because the length of some sub questions is shorter than 4.}
    \label{tab:bleu}
\end{table}
\begin{table}
    \centering
    \resizebox{\linewidth}{!}{
    \begin{tabular}{l|c|ccc}
    \hline
    Model & All  & Y/N     & Num   & Other    \\
    \hline
    Full  & \textbf{67.26} & \textbf{84.71} & \textbf{50.38}  & \textbf{58.44} \\
    wo-sub-loss & 66.94          & 84.58          & 48.95          & 58.29 \\
    wo-SQS & 66.55          & 84.43          & 46.78          & 58.18          \\
    \hline
    \end{tabular}}
    \caption{Ablation studies on impact of SQS on VQA 2.0 validation set.}
    \label{tab:vqa_aba1}
\end{table}
\subsection{Ablation Study}
\noindent
We conduct several ablation studies to explore critical factors affecting the performance of Co-VQA.
\paragraph{The impact of SQS.} In general, as we can observe from Table~\ref{tab:vqa_aba1}, though there are noises in the answers for SQs, the weak supervision signal provided by them shows a gain of +0.32. Furthermore, the decrease is obvious(-0.71) when we remove total SQS from the model, indicating that though the SQS generated from Questioner is not good enough, it still plays an important role in improving the performance of the model.
\paragraph{Detail Analysis of SQS.} To analyze the impact of SQS in detail, we divide the validation split of VQA-SQS into SQS-0 / SQS-1 / SQS-2 / SQS-3\&4 subsets, where SQS-n represents samples with n SQs. Then, the average accuracy of different models on each subset is reported in Table~\ref{tab:vqa_aba2}. For SQS-1 and SQS-2, the additional reasoning brought by SQS achieves an improvement of 1.02 and 0.93 respectively. However, for SQS-3\&4, the performance decreases compared with wo-SQS.
\begin{table}
    \centering
    \resizebox{\linewidth}{!}{
    \begin{tabular}{l|cccc|c}
    \hline
    Model & SQS-0 & SQS-1 & SQS-2  & SQS-3\&4 & All \\
     & (57,411) & (119,285) & (34,226) & (3,432) & (214,254) \\
    \hline
    Full& \textbf{69.51} & \textbf{66.70} & \textbf{65.78} & 63.62 & \textbf{67.26}\\
    wo-SQS & 69.48 & 65.68   & 64.85  & \textbf{64.35} & 66.55   \\
    \hline
    \end{tabular}}
    \caption{Ablation studies of SQS in detail on VQA 2.0 validation set. SQS-n represents the subset of samples with n SQs in VQA-SQS validation set. We report the average accuracy on each subset.}
    \label{tab:vqa_aba2}
\end{table}

\begin{table}
    \centering
    \resizebox{\linewidth}{!}{
    \begin{tabular}{l|cccc|c}
    \hline
    Subset & SQS-0 & SQS-1 & SQS-2  & SQS-3\&4 & All \\
    \hline
    Samples-Num & 57,411 & 119,285 & 34,226 & 3,432 & 214,254 \\
    Avg(Freq of SQ) & - & 870 & 851 & 693 & 854 \\
    \hline
    \end{tabular}}
    \caption{Data statistics of SQS in detail on VQA 2.0 validation set. The first row shows the number of original questions contained in different SQS sets, and the second row counts the average number of occurrences of the sub questions contained in each subset in the VQA-SQS train split.}
    \label{tab:vqa_aba22}
\end{table}

\begin{figure*}[t]
    \begin{center}
    \includegraphics[width=0.9\linewidth]{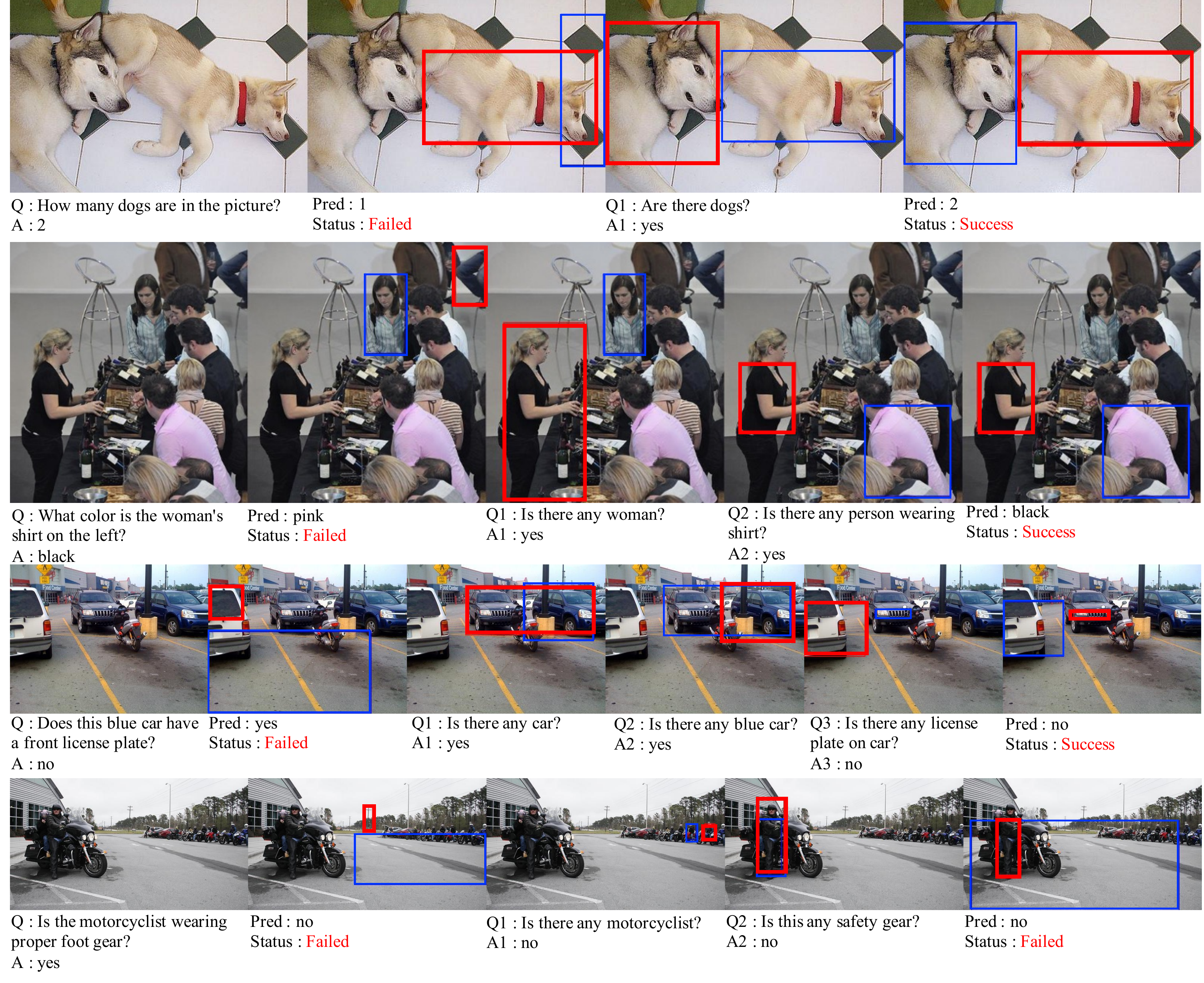}
    \end{center}
    \caption{Visualization of attention maps learned by complete Co-VQA with those learned by wo-SQS. The second and last column corresponds to the prediction of wo-SQS and complete Co-VQA respectively. Red and blue bounding boxes shown in each image are the top-2 attended regions.}
    \label{fig:case_study}
\end{figure*}

We perform statistics in two aspects to comprehensively explore the causes of this phenomenon. As shown in Table~\ref{tab:vqa_aba22}, compared with other subsets, SQS-3\&4 has fewer samples, causing insufficient learning for these samples of a long sequence. Moreover, SQs in SQS-3\&4 occur less frequently, thus it is inadequate for the model to establish accurate semantic connections between these images and questions.

\begin{table}
    \centering
    \resizebox{\linewidth}{!}{
    \begin{tabular}{l|c|ccc}
    \hline
    Model & All  & Y/N     & Num   & Other    \\
    \hline
    Full   & \textbf{67.26} & \textbf{84.71} & \textbf{50.38}  & \textbf{58.44} \\
    shuffle  & 67.15      & 84.68          & 49.78          & 58.42       \\
    random  & 67.08          & 84.68          & 49.75          & 58.28      \\
    \hline
    \end{tabular}}
    \caption{Ablation studies of coherence of SQS on VQA 2.0 validation set.}
    \label{tab:vqa_aba3}
\end{table}

\paragraph{Coherence of SQS.} We also study the impact of the coherence of SQS on performance. We run two different cases: 1) randomly shuffle the SQs in a sequence; 2) remove some SQs in a sequence with 50\% probability. As we can observe from Table~\ref{tab:vqa_aba3}, the declines from the original one are not significant, partly due to the fact that the coherence of SQS in the current dataset VQA-SQS is not good enough.

\subsection{Visualization} \label{section:vis}
\noindent
To better illustrate the effectiveness, explicit interpretability, and traceability of errors of Co-VQA, we visualize and compare the attention maps learned by complete Co-VQA with those learned by model wo-SQS. As shown in Figure~\ref{fig:case_study}. Column 1 is the original question and ground truth, while Column 2 corresponds to the prediction of model wo-SQS. The middle columns and last column correspond to the generated sub q\&a, and the prediction of Co-VQA, respectively. To visualize the attention maps, we use the in-degree of each node as the attention value and circle the top-2 attended regions with red and blue boxes.

Line 1 shows model wo-SQS only notices one of the dogs and gives a wrong answer "1". However, through SQ "Are there dogs?", Co-VQA focuses on two dogs and gives the correct answer "2". This case demonstrates that asking an existence question firstly is beneficial to number questions. In Line 2, model wo-SQS focuses on unrelated entities. However, Co-VQA attends to the women and the people wearing short sleeves gradually with SQS, and finally, concentrates on the related woman's shirt. Line 3 shows Co-VQA successively attends to cars, blue cars, and the license plate under the guidance of SQS and gets the correct answer. These examples prove that questions with different complexity will correspond to SQS of variable length, and SQ is indeed related to more accurate image attention. Moreover, generating SQ provides not only the logic of reasoning but also additional language interpretation. Thus, compared with previous works that only explain models by attention maps, Co-VQA has significantly better interpretability. 

The last line shows Co-VQA gives a wrong answer after adding SQS. However, we can find some possible causes, such as the wrong answer of Q1, Q2 is not related to the question, and the model doesn't attend to relevant entities in the light of Q1. It shows that Oracle and Questioner may give wrong answers or generate inappropriate questions, as well as Answerer may establish faulty semantic connections between questions and images, which verifies that Co-VQA has sure traceability for errors and provides guidance for future work.

\section{Conclusions}
\noindent
We propose a Conversation-based VQA (Co-VQA) framework which consists of Questioner, Oracle, and Answerer. Through internal conversation based on SQS, our model not only has explicit interpretability and traceability of answer errors but also can carry out question-adaptive variable-length reasoning chains. Currently, Questioner is relatively simple, and the quality still has a lot of room to improve. Meanwhile, current SQs are only yes/no questions. For future work, we plan to explore how to more effectively generate more diverse and higher quality SQS, and look forward to better model performance.

\section*{Acknowledgement}
\noindent
The work was partially supported by the National Natural Science Foundation of China (NSFC62076032) and the Cooperation Project with Beijing SanKuai Technology Co., Ltd. We would like to thank anonymous reviewers for their suggestions and comments, thank Duo Zhen for his suggestions about several iterations of this paper, thank our colleagues for their contributions in recorrecting the dataset.

\bibliography{anthology,custom}

\begin{thebibliography}{33}
\expandafter\ifx\csname natexlab\endcsname\relax\def\natexlab#1{#1}\fi

\bibitem[{Agrawal et~al.(2018)Agrawal, Batra, Parikh, and
  Kembhavi}]{Agrawal2018DontJA}
Aishwarya Agrawal, Dhruv Batra, Devi Parikh, and Aniruddha Kembhavi. 2018.
\newblock Don't just assume; look and answer: Overcoming priors for visual
  question answering.
\newblock \emph{2018 IEEE/CVF Conference on Computer Vision and Pattern
  Recognition}, pages 4971--4980.

\bibitem[{Agrawal et~al.(2015)Agrawal, Lu, Antol, Mitchell, Zitnick, Parikh,
  and Batra}]{Agrawal2015VQAVQ}
Aishwarya Agrawal, Jiasen Lu, Stanislaw Antol, Margaret Mitchell, C.~Lawrence
  Zitnick, Devi Parikh, and Dhruv Batra. 2015.
\newblock Vqa: Visual question answering.
\newblock \emph{International Journal of Computer Vision}, 123:4--31.

\bibitem[{Anderson et~al.(2018)Anderson, He, Buehler, Teney, Johnson, Gould,
  and Zhang}]{Anderson2018BottomUpAT}
Peter Anderson, Xiaodong He, Chris Buehler, Damien Teney, Mark Johnson, Stephen
  Gould, and Lei Zhang. 2018.
\newblock Bottom-up and top-down attention for image captioning and visual
  question answering.
\newblock \emph{2018 IEEE/CVF Conference on Computer Vision and Pattern
  Recognition}, pages 6077--6086.

\bibitem[{Ben-younes et~al.(2017)Ben-younes, Cad{\`e}ne, Cord, and
  Thome}]{Benyounes2017MUTANMT}
Hedi Ben-younes, R{\'e}mi Cad{\`e}ne, Matthieu Cord, and Nicolas Thome. 2017.
\newblock Mutan: Multimodal tucker fusion for visual question answering.
\newblock \emph{2017 IEEE International Conference on Computer Vision (ICCV)},
  pages 2631--2639.

\bibitem[{Cad{\`e}ne et~al.(2019)Cad{\`e}ne, Ben-younes, Cord, and
  Thome}]{Cadne2019MURELMR}
R{\'e}mi Cad{\`e}ne, Hedi Ben-younes, Matthieu Cord, and Nicolas Thome. 2019.
\newblock Murel: Multimodal relational reasoning for visual question answering.
\newblock \emph{2019 IEEE/CVF Conference on Computer Vision and Pattern
  Recognition (CVPR)}, pages 1989--1998.

\bibitem[{Chattopadhyay et~al.(2017)Chattopadhyay, Yadav, Prabhu,
  Chandrasekaran, Das, Lee, Batra, and Parikh}]{Chattopadhyay2017EvaluatingVC}
Prithvijit Chattopadhyay, Deshraj Yadav, Viraj Prabhu, Arjun Chandrasekaran,
  Abhishek Das, Stefan Lee, Dhruv Batra, and Devi Parikh. 2017.
\newblock Evaluating visual conversational agents via cooperative human-ai
  games.
\newblock In \emph{HCOMP}.

\bibitem[{Das et~al.(2017)Das, Kottur, Gupta, Singh, Yadav, Moura, Parikh, and
  Batra}]{Das2017VisualD}
Abhishek Das, Satwik Kottur, Khushi Gupta, Avi Singh, Deshraj Yadav, Jos{\'e}
  M.~F. Moura, Devi Parikh, and Dhruv Batra. 2017.
\newblock Visual dialog.
\newblock \emph{2017 IEEE Conference on Computer Vision and Pattern Recognition
  (CVPR)}, pages 1080--1089.

\bibitem[{de~Vries et~al.(2017)de~Vries, Strub, Chandar, Pietquin, Larochelle,
  and Courville}]{Vries2017GuessWhatVO}
Harm de~Vries, Florian Strub, A.~P.~Sarath Chandar, Olivier Pietquin,
  H.~Larochelle, and Aaron~C. Courville. 2017.
\newblock Guesswhat?! visual object discovery through multi-modal dialogue.
\newblock \emph{2017 IEEE Conference on Computer Vision and Pattern Recognition
  (CVPR)}, pages 4466--4475.

\bibitem[{Gao et~al.(2019)Gao, Li, You, Jiang, Lu, Hoi, and
  Wang}]{Gao2019DynamicFW}
Peng Gao, Hongsheng Li, Haoxuan You, Zhengkai Jiang, Pan Lu, Steven C.~H. Hoi,
  and Xiaogang Wang. 2019.
\newblock Dynamic fusion with intra- and inter-modality attention flow for
  visual question answering.
\newblock \emph{2019 IEEE/CVF Conference on Computer Vision and Pattern
  Recognition (CVPR)}, pages 6632--6641.

\bibitem[{Goyal et~al.(2017{\natexlab{a}})Goyal, Doll{\'a}r, Girshick,
  Noordhuis, Wesolowski, Kyrola, Tulloch, Jia, and He}]{Goyal2017AccurateLM}
Priya Goyal, Piotr Doll{\'a}r, Ross~B. Girshick, Pieter Noordhuis, Lukasz
  Wesolowski, Aapo Kyrola, Andrew Tulloch, Yangqing Jia, and Kaiming He.
  2017{\natexlab{a}}.
\newblock Accurate, large minibatch sgd: Training imagenet in 1 hour.
\newblock \emph{ArXiv}, abs/1706.02677.

\bibitem[{Goyal et~al.(2017{\natexlab{b}})Goyal, Khot, Summers-Stay, Batra, and
  Parikh}]{Goyal2017MakingTV}
Yash Goyal, Tejas Khot, Douglas Summers-Stay, Dhruv Batra, and Devi Parikh.
  2017{\natexlab{b}}.
\newblock Making the v in vqa matter: Elevating the role of image understanding
  in visual question answering.
\newblock \emph{2017 IEEE Conference on Computer Vision and Pattern Recognition
  (CVPR)}, pages 6325--6334.

\bibitem[{Hu et~al.(2018)Hu, Gu, Zhang, Dai, and Wei}]{Hu2018RelationNF}
Han Hu, Jiayuan Gu, Zheng Zhang, Jifeng Dai, and Yichen Wei. 2018.
\newblock Relation networks for object detection.
\newblock \emph{2018 IEEE/CVF Conference on Computer Vision and Pattern
  Recognition}, pages 3588--3597.

\bibitem[{Jiang et~al.(2018)Jiang, Natarajan, Chen, Rohrbach, Batra, and
  Parikh}]{Jiang2018PythiaVT}
Yu~Jiang, Vivek Natarajan, Xinlei Chen, Marcus Rohrbach, Dhruv Batra, and Devi
  Parikh. 2018.
\newblock Pythia v0.1: the winning entry to the vqa challenge 2018.
\newblock \emph{ArXiv}, abs/1807.09956.

\bibitem[{Kim et~al.(2018)Kim, Jun, and Zhang}]{Kim2018BilinearAN}
Jin-Hwa Kim, Jaehyun Jun, and Byoung-Tak Zhang. 2018.
\newblock Bilinear attention networks.
\newblock In \emph{NeurIPS}.

\bibitem[{Krishna et~al.(2017)Krishna, Zhu, Groth, Johnson, Hata, Kravitz,
  Chen, Kalantidis, Li, Shamma et~al.}]{krishna2017visual}
Ranjay Krishna, Yuke Zhu, Oliver Groth, Justin Johnson, Kenji Hata, Joshua
  Kravitz, Stephanie Chen, Yannis Kalantidis, Li-Jia Li, David~A Shamma, et~al.
  2017.
\newblock Visual genome: Connecting language and vision using crowdsourced
  dense image annotations.
\newblock \emph{International journal of computer vision}, 123(1):32--73.

\bibitem[{Leslie(1987)}]{Leslie1987PretenseAR}
Alan~M. Leslie. 1987.
\newblock Pretense and representation: The origins of "theory of mind.".
\newblock \emph{Psychological Review}, 94:412--426.

\bibitem[{Li et~al.(2019)Li, Gan, Cheng, and Liu}]{Li2019RelationAwareGA}
Linjie Li, Zhe Gan, Yu~Cheng, and Jingjing Liu. 2019.
\newblock Relation-aware graph attention network for visual question answering.
\newblock \emph{2019 IEEE/CVF International Conference on Computer Vision
  (ICCV)}, pages 10312--10321.

\bibitem[{Lin et~al.(2014)Lin, Maire, Belongie, Hays, Perona, Ramanan,
  Doll{\'a}r, and Zitnick}]{Lin2014MicrosoftCC}
Tsung-Yi Lin, Michael Maire, Serge~J. Belongie, James Hays, Pietro Perona, Deva
  Ramanan, Piotr Doll{\'a}r, and C.~Lawrence Zitnick. 2014.
\newblock Microsoft coco: Common objects in context.
\newblock In \emph{ECCV}.

\bibitem[{Loper and Bird(2002)}]{journals/corr/cs-CL-0205028}
Edward Loper and Steven Bird. 2002.
\newblock \href
  {http://dblp.uni-trier.de/db/journals/corr/corr0205.html#cs-CL-0205028}
  {Nltk: The natural language toolkit}.
\newblock \emph{CoRR}, cs.CL/0205028.

\bibitem[{Lu et~al.(2016)Lu, Yang, Batra, and Parikh}]{Lu2016HierarchicalQC}
Jiasen Lu, Jianwei Yang, Dhruv Batra, and Devi Parikh. 2016.
\newblock Hierarchical question-image co-attention for visual question
  answering.
\newblock In \emph{NIPS}.

\bibitem[{Malinowski et~al.(2018)Malinowski, Doersch, Santoro, and
  Battaglia}]{Malinowski2018LearningVQ}
Mateusz Malinowski, Carl Doersch, Adam Santoro, and Peter~W. Battaglia. 2018.
\newblock Learning visual question answering by bootstrapping hard attention.
\newblock In \emph{ECCV}.

\bibitem[{Nam et~al.(2017)Nam, Ha, and Kim}]{Nam2017DualAN}
Hyeonseob Nam, Jung-Woo Ha, and Jeonghee Kim. 2017.
\newblock Dual attention networks for multimodal reasoning and matching.
\newblock \emph{2017 IEEE Conference on Computer Vision and Pattern Recognition
  (CVPR)}, pages 2156--2164.

\bibitem[{Nguyen and Okatani(2018)}]{Nguyen2018ImprovedFO}
Duy-Kien Nguyen and Takayuki Okatani. 2018.
\newblock Improved fusion of visual and language representations by dense
  symmetric co-attention for visual question answering.
\newblock \emph{2018 IEEE/CVF Conference on Computer Vision and Pattern
  Recognition}, pages 6087--6096.

\bibitem[{Paszke et~al.(2017)Paszke, Gross, Chintala, Chanan, Yang, DeVito,
  Lin, Desmaison, Antiga, and Lerer}]{Paszke2017AutomaticDI}
Adam Paszke, Sam Gross, Soumith Chintala, Gregory Chanan, Edward Yang, Zach
  DeVito, Zeming Lin, Alban Desmaison, Luca Antiga, and Adam Lerer. 2017.
\newblock Automatic differentiation in pytorch.

\bibitem[{Ren et~al.(2015)Ren, He, Girshick, and Sun}]{Ren2015FasterRT}
Shaoqing Ren, Kaiming He, Ross~B. Girshick, and Jian Sun. 2015.
\newblock Faster r-cnn: Towards real-time object detection with region proposal
  networks.
\newblock \emph{IEEE Transactions on Pattern Analysis and Machine
  Intelligence}, 39:1137--1149.

\bibitem[{Saha et~al.(2018)Saha, Khapra, and
  Sankaranarayanan}]{Saha2018TowardsBL}
Amrita Saha, Mitesh~M. Khapra, and Karthik Sankaranarayanan. 2018.
\newblock Towards building large scale multimodal domain-aware conversation
  systems.
\newblock In \emph{AAAI}.

\bibitem[{Selvaraju et~al.(2020)Selvaraju, Tendulkar, Parikh, Horvitz, Ribeiro,
  Nushi, and Kamar}]{Selvaraju2020SQuINTingAV}
Ramprasaath~R. Selvaraju, Purva Tendulkar, Devi Parikh, Eric Horvitz,
  Marco~T{\'u}lio Ribeiro, Besmira Nushi, and Ece Kamar. 2020.
\newblock Squinting at vqa models: Introspecting vqa models with sub-questions.
\newblock \emph{2020 IEEE/CVF Conference on Computer Vision and Pattern
  Recognition (CVPR)}, pages 10000--10008.

\bibitem[{Sordoni et~al.(2015)Sordoni, Bengio, Vahabi, Lioma, Simonsen, and
  Nie}]{Sordoni2015AHR}
Alessandro Sordoni, Yoshua Bengio, Hossein Vahabi, Christina Lioma, Jakob~Grue
  Simonsen, and Jianyun Nie. 2015.
\newblock A hierarchical recurrent encoder-decoder for generative context-aware
  query suggestion.
\newblock \emph{Proceedings of the 24th ACM International on Conference on
  Information and Knowledge Management}.

\bibitem[{Vaswani et~al.(2017)Vaswani, Shazeer, Parmar, Uszkoreit, Jones,
  Gomez, Kaiser, and Polosukhin}]{Vaswani2017AttentionIA}
Ashish Vaswani, Noam~M. Shazeer, Niki Parmar, Jakob Uszkoreit, Llion Jones,
  Aidan~N. Gomez, Lukasz Kaiser, and Illia Polosukhin. 2017.
\newblock Attention is all you need.
\newblock \emph{ArXiv}, abs/1706.03762.

\bibitem[{Velickovic et~al.(2018)Velickovic, Cucurull, Casanova, Romero,
  Lio’, and Bengio}]{Velickovic2018GraphAN}
Petar Velickovic, Guillem Cucurull, Arantxa Casanova, Adriana Romero, Pietro
  Lio’, and Yoshua Bengio. 2018.
\newblock Graph attention networks.
\newblock \emph{ArXiv}, abs/1710.10903.

\bibitem[{Yang et~al.(2016)Yang, He, Gao, Deng, and Smola}]{Yang2016StackedAN}
Zichao Yang, Xiaodong He, Jianfeng Gao, Li~Deng, and Alex Smola. 2016.
\newblock Stacked attention networks for image question answering.
\newblock \emph{2016 IEEE Conference on Computer Vision and Pattern Recognition
  (CVPR)}, pages 21--29.

\bibitem[{Yu et~al.(2019)Yu, Yu, Cui, Tao, and Tian}]{Yu2019DeepMC}
Zhou Yu, Jun Yu, Yuhao Cui, Dacheng Tao, and Qi~Tian. 2019.
\newblock Deep modular co-attention networks for visual question answering.
\newblock \emph{2019 IEEE/CVF Conference on Computer Vision and Pattern
  Recognition (CVPR)}, pages 6274--6283.

\bibitem[{Yu et~al.(2017)Yu, Yu, Fan, and Tao}]{Yu2017MultimodalFB}
Zhou Yu, Jun Yu, Jianping Fan, and Dacheng Tao. 2017.
\newblock Multi-modal factorized bilinear pooling with co-attention learning
  for visual question answering.
\newblock \emph{2017 IEEE International Conference on Computer Vision (ICCV)},
  pages 1839--1848.

\end{thebibliography}
\bibliographystyle{acl_natbib}

\appendix
\clearpage
\section{Appendix}
\noindent
Here we introduce our method for constructing SQS and the statistical information of datasets.
\subsection{Data source} 
\noindent
We construct our SQS dataset based on VQA 2.0 and VQA-CP v2 datasets. 
\subsection{Construction principle}
\noindent
To accomplish the process of SQS construction, we first determine the order of questions according to the templates in Table~\ref{tab:question-order}. For order-0 and order-1 questions, there is no corresponding SQ, order-2 questions can construct the corresponding order-1 SQs, while the order-3 questions can construct multiple order-1 SQs and order-2 SQs. Then, the principle of dataset construction is: high-order questions can adopt corresponding low-order questions as their sub questions, for each high-order question, these sub questions are arranged according to the order from low to high to form a sub question sequence.

\subsection{Construction method}
\noindent
The details of the construction method can be illustrated as following:

\noindent
1) For each question, we first adopt Spacy\textsuperscript{1} \footnote{\textsuperscript{1}https://spacy.io/} and NLTK toolkit \cite{journals/corr/cs-CL-0205028} to identify all noun blocks in the question and filter out some noun blocks based on the predefined phrase list. The phrase list mainly includes meaningless quantifiers, pronouns, and abstract nouns, such as lots, someone, something, you, they, it, the day, the picture, a body, emotion, this, type, etc. \\
2) After finishing the filter process, for questions that still contain noun blocks, according to the dependency relation between the extracted noun blocks, part of these noun blocks may be used as prepositional phrases. For other remaining noun blocks, we use Part-of-Speech Tagging of Spacy to classify them into corresponding nouns, adjectives, quantifiers, and prepositional phrases. For nouns, we save them separately, while for adjectives, quantifiers, and prepositional phrases, we save these modifiers together with the noun blocks in a form of 2-tuple (noun, modifier), such as (flower, red).

\begin{table}
    \centering
    \resizebox{\linewidth}{!}{
    \begin{tabular}{l|c}
    \hline
    order &  question template  \\
    \hline
    0  &  no entity \\
    1  &  single entity\\
    2  &  entity \& attribute \\
    3  &  comparison between different entities \\
    \hline
    \end{tabular}}
    \vspace{0.1cm}\caption{Templates for question of different order.}
    \label{tab:question-order}
\end{table}
\begin{table*}[htp]
    \begin{center}
    \begin{tabular}{l|c}
    \hline
    Question Type & Matching Pattern \\
    \hline
    Existence & (do you see)?[DET | PRON | ADP]* [NOUN | PROPN]* NOUN? \\
    \hline
    Verb & (do you see)? [DET | PRON | ADP]* [NOUN | PROPN]* NOUN? \\
      & [VBG | VBN]? \\
    \hline
    Attribute & BE [DET | PRON | ADP]* [NOUN | PROPN]* NOUN? ADJ? \\
    \hline
    Num & BE [DET | PRON | ADP]* NUM NOUN NOUN* ? \\
    \hline
    Prep & BE [DET | PRON | ADP]* [NOUN | PROPN]* NOUN VERB? ADP DET \\
     &  NOUN NOUN* ?\\
    \hline
    \end{tabular}
    \vspace{0.1cm}\caption{Matching patterns for different type of questions}
    \label{tab:Match-pattern}
    \end{center}
\end{table*}
\begin{table*}
    \centering
    \resizebox{\linewidth}{!}{
    \begin{tabular}{cc|cccc}
    \hline
    Dataset & Split & \#Images & \#Q\&A & \#Non-empty SQS & Avg(\#SQ) \\
    \hline
    VQA-SQS & Train & 82,783 & 443,757 & 328,140 & 0.94 \\
    VQA-SQS & Val & 40,504 & 214,354 & 156,943 & 0.925\\
    VQA-CP-SQS & Train & 120,932 & 438,183 & 322,200 & 0.93 \\
    VQA-CP-SQS & Test & 98,226 & 219,928 & 162,883 & 0.946 \\
    \hline
    \end{tabular}}
    \vspace{0.1cm}\caption{Dataset statistics of VQA-SQS and VQA-CP-SQS.}
    \label{tab:dataset}
\end{table*}
\begin{table}[htp]
    \newcommand{\tabincell}[2]{\begin{tabular}{@{}#1@{}}#2\end{tabular}}%
    \begin{center}
    \begin{tabular}{l|c}
    \hline
    SQ Type & Matching Pattern \\
    \hline
    Existence & \tabincell{c}{Is there any [entity]? \\ Is there any [color] [entity]? \\ Are there [entites]?} \\
    \hline
    Attribute & \tabincell{c}{Is the [entity] [color]? \\ Is any [entity]? \\ Are these [entites] in similar size?} \\
    \hline
    Prep & \tabincell{c}{Is there any [entity] on the [entity2]? \\ Is there any [entity] in the [entity2]?} \\
    \hline
    Number & \tabincell{c}{Are there [number] [entites]? \\ Is there only one [entity]?} \\
    \hline
    Position & \tabincell{c}{Is the [entity] on the left? \\ Is the [entity] on the right? \\ Is the [entity] in the middle?} \\
    \hline
    \end{tabular}
    \vspace{0.1cm}\caption{Sub question generation template for different SQ types.}
    \label{tab:sub-Match-pattern}
    \end{center}
\end{table}

\noindent
3) After step 1, for questions without noun blocks, considering there may be omissions in the process of extraction, we perform pattern matching through Spacy based on the pre-defined matching template to determine the category of these questions. Table~\ref{tab:Match-pattern} illustrates partial matching patterns for different type of questions.
Especially, for existence questions, no additional processing is required, while for other types of questions, we save the nouns that are existing in the questions. \\
\noindent
4) We further filter the nouns and tuples saved in 2) and 3). This conduction aims to filter out abstract nouns, non-substantial nouns, and 2-tuple corresponding to these nouns. The following are some cases to be filtered: \\
a) \textbf{Abstract Noun}:  direction, design, surface, area, emotion, skill etc. \\
b) \textbf{Non Substantive Noun}: mode, base, day, love, name, print, piece etc. \\
\begin{figure*}
    \begin{center}
    \includegraphics[width=1.0\linewidth]{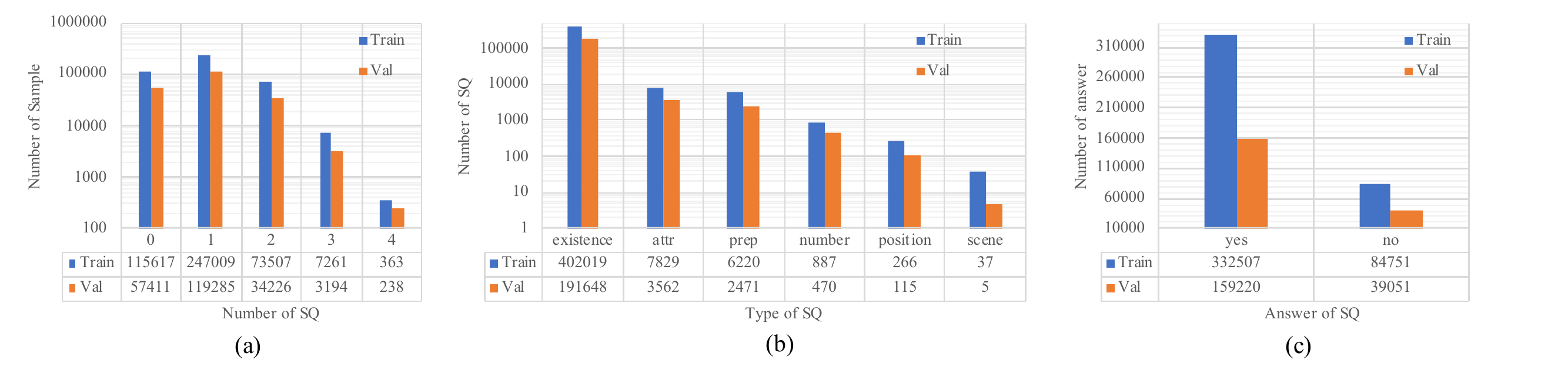}
    \end{center}
    \caption{Dataset distribution of VQA-SQS.}
    \label{fig:VQA-SQS}
\end{figure*}
\begin{figure*}
    \begin{center}
    \includegraphics[width=1.0\linewidth]{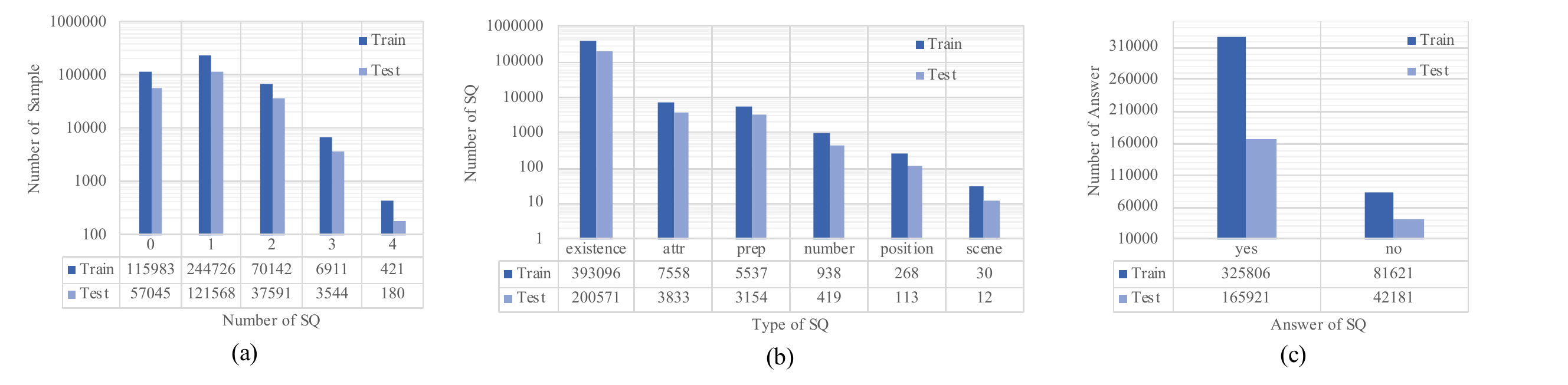}
    \end{center}
    \caption{Dataset distribution of VQA-CP-SQS.}
    \label{fig:CP-SQS}
\end{figure*}
\noindent
5) For the remaining nouns and their corresponding 2-tuple, we use the pre-defined question template to construct the corresponding sub questions. To facilitate the process of construction, we design all sub questions as yes / no questions.  The matching pattern for each type of sub question are revealed in Table~\ref{tab:sub-Match-pattern}. \\
6) The construction process of ground-truth answers for sub questions can be illustrated as follows: \\
\textbf{Existence SQ and Attribute SQ}  we first extract the label and attribute information of the entity by using the detection model and then combine this information to produce the answer. \\
\begin{figure*}
    \begin{center}
    \includegraphics[width=1.0\linewidth]{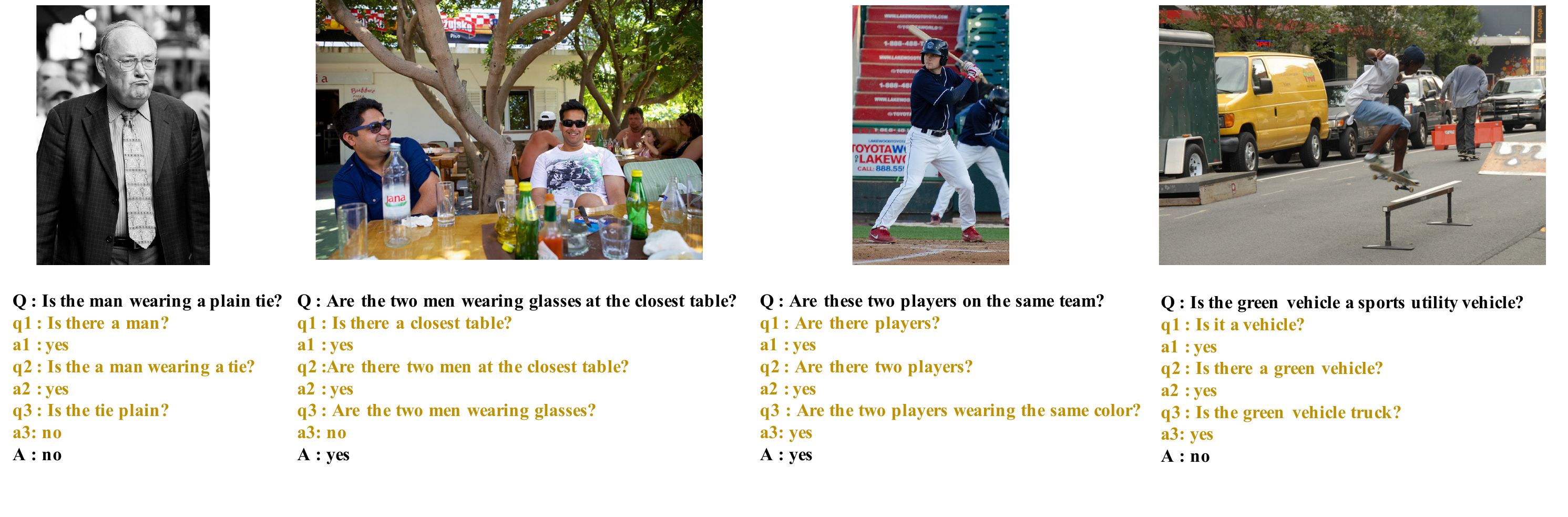}
    \end{center}
    \caption{Some samples of VQA-SQS, including existence SQ, attribute SQ, prep SQ and number SQ.}
    \label{fig:SQS-case}
\end{figure*}
\noindent
\textbf{Prep SQ and Position SQ}  the location information obtained by the detection model is utilized to judge the relationship of overlapping and orientation between entities, we use the obtained relationship to generate the corresponding answer. \\
\textbf{Number SQ} we first make a rough quantity estimation based on the image, and then make a manual correction.  \\
\noindent
6) Considering there may be wrong answers, incoherent sequences, and nonstandard question grammar in the process of automatic construction. So, to increase the diversity of SQs, we invite ten students in our laboratory to further manually recorrect some samples(about 5K samples). \\
The SQS datasets obtained by performing the above operations on VQA 2.0 and VQA-CP v2 datasets are called VQA-SQS and VQA-CP-SQS respectively.

\subsection{Dataset statistics}
\noindent
Table~\ref{tab:dataset} shows general statistical information of the two SQS datasets, then, Figure~\ref{fig:VQA-SQS} and Figure~\ref{fig:CP-SQS} respectively reveal three fine-grained distributions of two datasets including number distribution of SQ (7-a / 8-a), type distribution of SQ (7-b / 8-b) and answer distribution of SQ (7-c / 8-c). To display more convenient, in (7-a / 8-a) and (7-b / 8-b), the ordinate axis adopts logarithmic scale.

Figure~\ref{fig:SQS-case} displays four samples with three sub questions in the VQA-SQS dataset.

\end{document}